\def\year{2017}\relax
\documentclass[journal,transmag]{IEEEtran}

\usepackage{times}
\usepackage{helvet}
\usepackage{courier}

\usepackage{epsfig}
\usepackage{graphicx}
\usepackage{amsmath}
\usepackage{amssymb}
\usepackage{bm}
\usepackage{mysymbols}
\usepackage{caption}
\usepackage{subcaption}
\usepackage{float}

\usepackage{multirow}
\usepackage{makecell}

\begin{document}

\title{Semantic Jitter:\\Dense Supervision for Visual Comparisons via Synthetic Images}

\author{Aron Yu\\
University of Texas at Austin\\
{\tt\small aron.yu@utexas.edu}
\and
Kristen Grauman\\
University of Texas at Austin\\
{\tt\small grauman@cs.utexas.edu}
}

\maketitle

\begin{abstract}
  Distinguishing subtle differences in attributes is valuable, yet learning to make visual comparisons remains non-trivial.   Not only is the number of possible comparisons quadratic in the number of training images, but also access to images adequately spanning the space of fine-grained visual differences is limited.  We propose to overcome the \emph{sparsity of supervision} problem  via synthetically generated images.  Building on a state-of-the-art image generation engine, we sample pairs of training images exhibiting slight modifications of individual attributes. Augmenting real training image pairs with these examples, we then train attribute ranking models to predict the relative strength of an attribute in novel pairs of real images.  Our results on datasets of faces and fashion images show the great promise of bootstrapping imperfect image generators to counteract sample sparsity for learning to rank.
\end{abstract}

\section{Introduction}
\label{sec:intro}
\vspace*{0.05in}

Fine-grained analysis of images often entails making \emph{visual comparisons}.  For example, given two products in a fashion catalog, a shopper may judge which shoe appears more pointy at the toe.  Given two selfies, a teen may gauge in which one he is smiling more.  Given two photos of houses for sale on a real estate website, a home buyer may analyze which facade looks better maintained.  Given a series of MRI scans, a radiologist may judge which pair exhibits the most shape changes.

In these and many other such cases, we are interested in inferring how a pair of images compares in terms of a particular property, or ``attribute''.  That is, which is more \emph{pointy}, \emph{smiling}, \emph{well-maintained}, etc.  Importantly, the distinctions of interest are often quite subtle.  Subtle comparisons arise both in image pairs that are very similar in almost every regard (e.g., two photos of the same individual wearing the same clothing, yet smiling more in one photo than the other), as well as image pairs that are holistically different yet exhibit only slight differences in the attribute in question (e.g., two individuals different in appearance, and one is smiling slightly more than the other).


\begin{figure}[t]
  \captionsetup{font=footnotesize}
  \centering
  \includegraphics[width=0.46\textwidth]{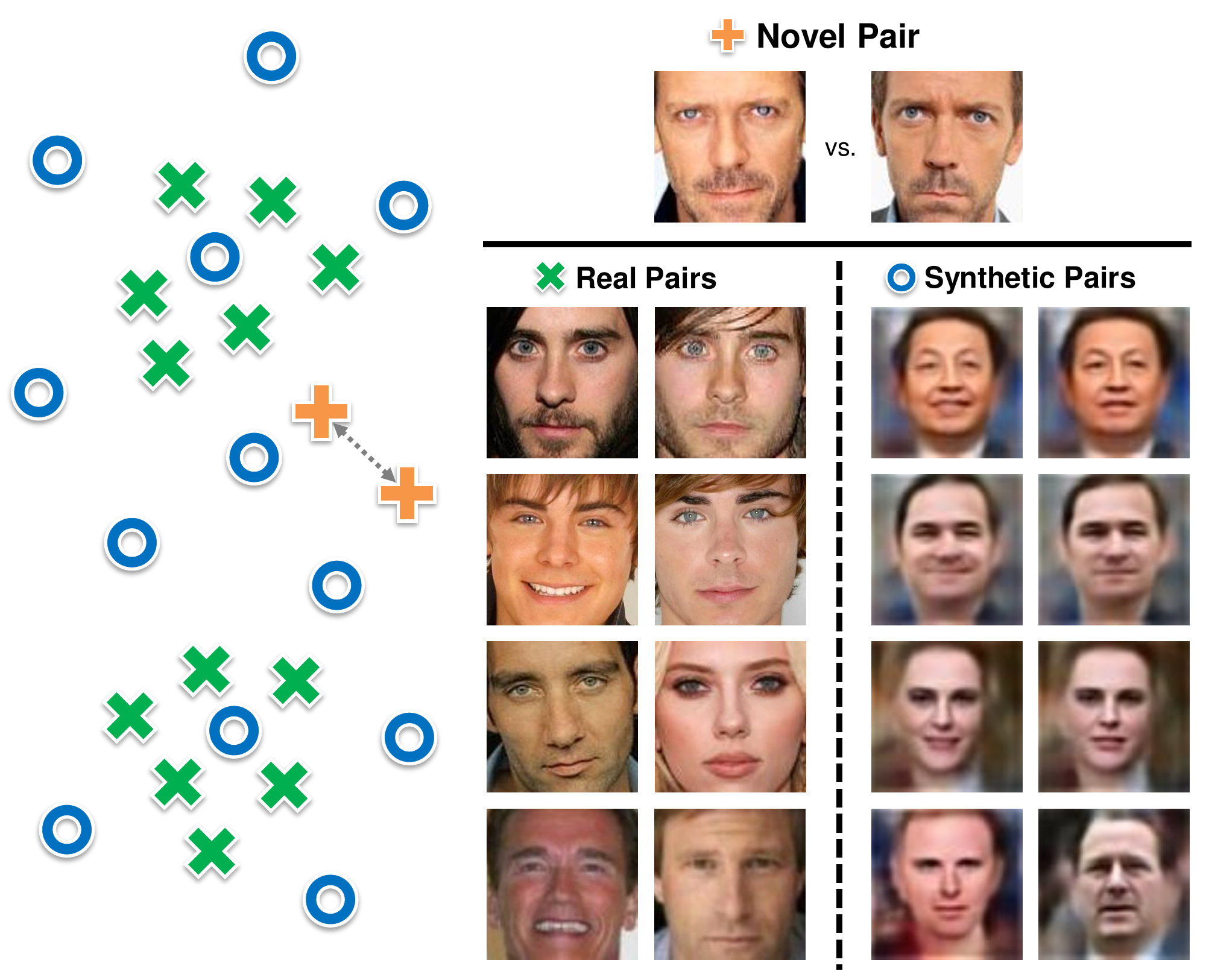}
  \caption{Our method ``densifies'' supervision for training ranking functions to make visual comparisons, by generating ordered pairs of synthetic images.  Here, when learning the attribute \emph{smiling}, real training images need not be representative of the entire attribute space (e.g., Web photos may cluster around commonly photographed expressions, like toothy smiles).  Our idea ``fills in'' the sparsely sampled regions to enable fine-grained supervision.  Given a novel pair (top), the nearest synthetic pairs (right) may present better training data than the nearest real pairs (left).}
  \label{fig:concept}
\end{figure}

A growing body of work explores computational models for visual comparisons~\cite{Datta11,whittle-ijcv,Li12,relative-attributes,Sandeep14,gupta-eccv2012,Singh16,Souri16,Xiao15,Yang16,aron-cvpr2014,aron-iccv2015}.   In particular, ranking models for ``relative attributes''~\cite{whittle-ijcv,Li12,relative-attributes,Sandeep14,Xiao15,aron-cvpr2014} use human-ordered pairs of images to train a system to predict the relative ordering in novel image pairs.

A major challenge in training a ranking model is the \emph{sparsity of supervision}.  That sparsity stems from two factors: label availability and image availability.  Because training instances consist of pairs of images---together with the ground truth human judgment about which exhibits the property more or less---the space of all possible comparisons is quadratic in the number of potential training images.  This quickly makes it intractable to  label an image collection exhaustively for its comparative properties.  At the same time, attribute comparisons entail a greater cognitive load than, for example, object category labeling.  Indeed, the largest existing relative attribute datasets sample only less than 0.1\% of all image pairs for ground truth labels~\cite{aron-cvpr2014}, and there is a major size gap between standard datasets labeled for classification (now in the millions~\cite{imagenet}) and those for comparisons (at best in the thousands~\cite{aron-cvpr2014}).  A popular shortcut is to propagate category-level comparisons down to image instances~\cite{relative-attributes,biswas-cvpr2013}---e.g., deem all ocean scenes as ``more open'' than all forest scenes---but this introduces substantial label noise and in practice underperforms training with instance-level comparisons~\cite{whittle-ijcv}.

Perhaps more insidious than the annotation cost, however, is the problem of even \emph{curating} training images that sufficiently illustrate fine-grained differences.  Critically, sparse supervision arises not simply because 1) we lack resources to get enough image pairs labeled, but also because 2) \emph{we lack a direct way to curate photos demonstrating all sorts of subtle attribute changes}.  For example, how might we gather unlabeled image pairs depicting all subtle differences in ``sportiness" in clothing images or ``surprisedness'' in faces?  As a result, even today's best datasets contain only partial representations of an attribute.  See Figure~\ref{fig:concept}.

We propose to use \emph{synthetic image pairs} to overcome the sparsity of supervision problem when learning to compare images.  The main idea is to synthesize plausible photos exhibiting variations along a given attribute from a generative model, thereby recovering samples in regions of the attribute space that are underrepresented among the real training images.  After (optionally) verifying the comparative labels with human annotators, we train a discriminative ranking model using the synthetic training pairs in conjunction with real image pairs.  The resulting model predicts attribute comparisons between novel pairs of real images.

\begin{figure}[t]
  \captionsetup{font=footnotesize}
  \centering
  \includegraphics[width=0.46\textwidth]{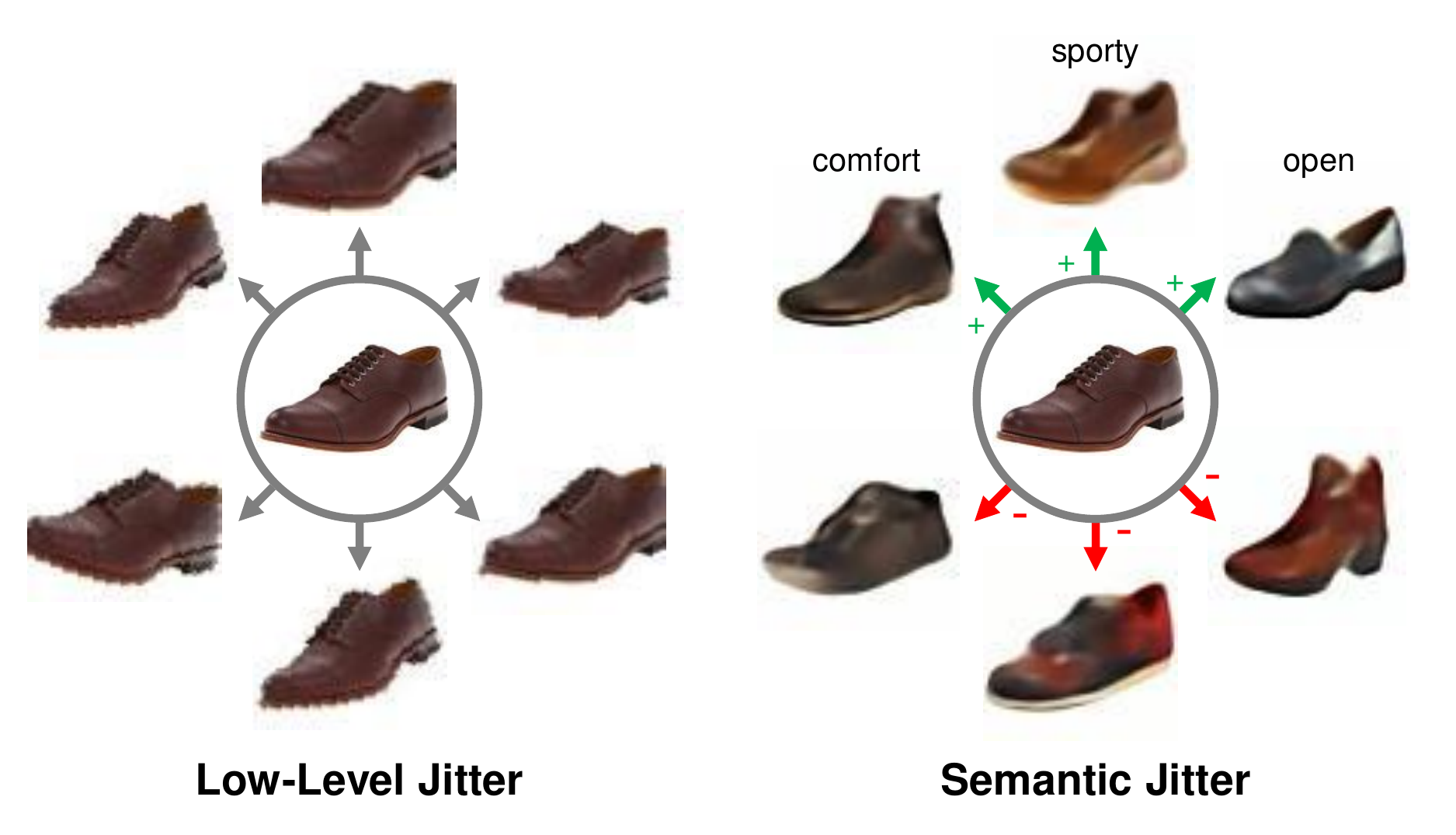}
  \vspace*{-0.08in}
  \caption{Whereas standard data augmentation with low-level ``jitter'' (left) expands training data with image-space alterations (mirroring, scaling, etc.), our \emph{semantic jitter} (right) expands training data with high-level alterations, tweaking semantic properties in a controlled manner.  Best viewed in color.}
  \vspace*{-0.15in}
  \label{fig:jitter_comp}
\end{figure}

Our idea can be seen as semantic ``jittering'' of the data to augment real image training sets with nearby variations.  The systemic perturbation of images through label-preserving transforms like mirroring/scaling is now common practice in training deep networks for classification~\cite{Dosovitskiy14,Simard03,Vincent08}.  Whereas such low-level image manipulations are performed independent of the semantic content of the training instance, the variations introduced by our approach are high-level changes that affect the very meaning of the image, e.g., facial shape changes as the expression changes.  In other words, our jitter has a semantic basis rather than a purely geometric/photometric basis.  See Figure~\ref{fig:jitter_comp}.

We demonstrate our approach in domains where subtle visual comparisons are often relevant: faces and fashion.  To support our experiments, we crowdsource a lexicon of fine-grained attributes that people naturally use to describe subtle differences, and we gather new comparison annotations.  In both domains---and for two distinct popular ranking models---we show that artificially ``densifying'' comparative supervision improves precise attribute predictions.

\section{Related Work}
\label{sec:related}
\vspace*{0.02in}

\noindent
\textbf{Attribute Comparisons:}  Since the introduction of relative attributes~\cite{relative-attributes}, the task of attribute comparisons has gained attention for its variety of applications, such as online shopping \cite{whittle-ijcv}, biometrics~\cite{nixon-attributes}, novel forms of low-supervision learning~\cite{gupta-eccv2012,biswas-cvpr2013}, and font selection~\cite{adobe-relative-attributes-fonts}.

The original approach~\cite{relative-attributes} adopts a learning-to-rank framework~\cite{Joachims}.  Pairwise supervision is used to train a linear ranking function for each attribute.  More recently, non-linear ranking functions~\cite{Li12}, combining feature-specific rankers~\cite{Datta11}, and training local rankers on the fly~\cite{aron-cvpr2014,aron-iccv2015} are all promising ways to improve accuracy.  Other work investigates features tailored for attribute comparisons, such as facial landmark detectors~\cite{Sandeep14} and \textit{visual chains} to discover relevant parts~\cite{Xiao15}.  The success of deep networks has motivated end-to-end frameworks for learning features and attribute ranking functions simultaneously~\cite{Singh16,Souri16,Yang16}.  Unlike any of the above, the novelty of our idea rests in the source data for training, not the learning algorithm.  We evaluate its benefits for two popular ranking frameworks---RankSVM~\cite{whittle-ijcv,relative-attributes,aron-cvpr2014,aron-iccv2015,biswas-cvpr2013,Joachims} and a Siamese deep convolutional neural network (CNN)~\cite{Singh16}.


\vspace*{0.1in}
\noindent
\textbf{Attributes and Image Synthesis:}  Our approach relies on a generative model for image synthesis that can progressively modify a target attribute.  Attribute-specific alterations have been considered in several recent methods, primarily for face images.  Some target a specific domain and attribute, such as methods to enhance the ``memorability''~\cite{Khosla13} or age~\cite{Kemelmacher14} of facial photos, or to edit outdoor scenes with transient attributes like weather~\cite{Laffont14}.  Alternatively, the success of deep neural networks for image generation (i.e., Generative Adversarial Nets (GAN)~\cite{Goodfellow14} or Variational Auto-Encoders (VAE)~\cite{Gregor15,Kingma14,Kulkarni15}) opens the door to learning how to generate images conditioned on desired properties~\cite{Dosovitskiy15,Li16,Pandey16,Yan16-eccv,Yan16-arxiv}.  For example, a conditional multimodal auto-encoder can generate faces from attribute descriptions~\cite{Pandey16}, and focus on identity-preserving changes~\cite{Li16}.  We employ the state-of-the-art model of~\cite{Yan16-eccv} due to its generality.  Whereas the above methods aim to produce an image for human inspection, we aim to generate dense supervision for learning algorithms.

\vspace*{0.1in}
\noindent
\textbf{Training Recognition Models with Synthetic Images:}  The use of synthetic images as training data has been explored to a limited extent, primarily for human bodies.  Taking advantage of high quality graphics models for humanoids, rendered images of people from various viewpoints and body poses provide free data to train pose estimators~\cite{psh,shotton-cvpr2011} or person detectors~\cite{pishchulin-bmvc2011}.  After manually marking a pose in the first frame of a video, one can personalize a pose estimator by synthesizing deformations~\cite{ParkRamanan}.  For objects beyond people, recent work considers how to exploit non-photorealistic images generated from 3D CAD models to augment training sets for object detection~\cite{saenko-iccv2015}, or words rendered in different fonts for text recognition~\cite{jaderberg-text-2014}.

While these methods share our concern about the sparsity of supervision, our focus on attributes and ranking is unique.  Furthermore, most  methods assume a graphics engine and 3D model to render new views with desired parameters (pose, viewpoint, etc.).  In contrast, we investigate images generated from a 2D image synthesis engine in which the modes of variation are controlled by a \emph{learned} model.  Being data-driven can offer greater flexibility, allowing tasks beyond those requiring a 3D model, and variability beyond camera pose and lighting parameters.

\section{Approach}
\label{sec:approach}
\vspace*{0.05in}

Our idea is to ``densify'' supervision for learning to make visual comparisons, by leveraging images sampled from an attribute-conditioned generative model.  First we overview the visual comparison task.  Then, we describe the generative model and how we elicit dense supervision pairs from it.  Finally, we integrate synthetic and real images to train rankers for attribute comparisons.


\begin{figure}[t]
  \captionsetup{font=footnotesize}
  \centering
  \includegraphics[width=0.48\textwidth]{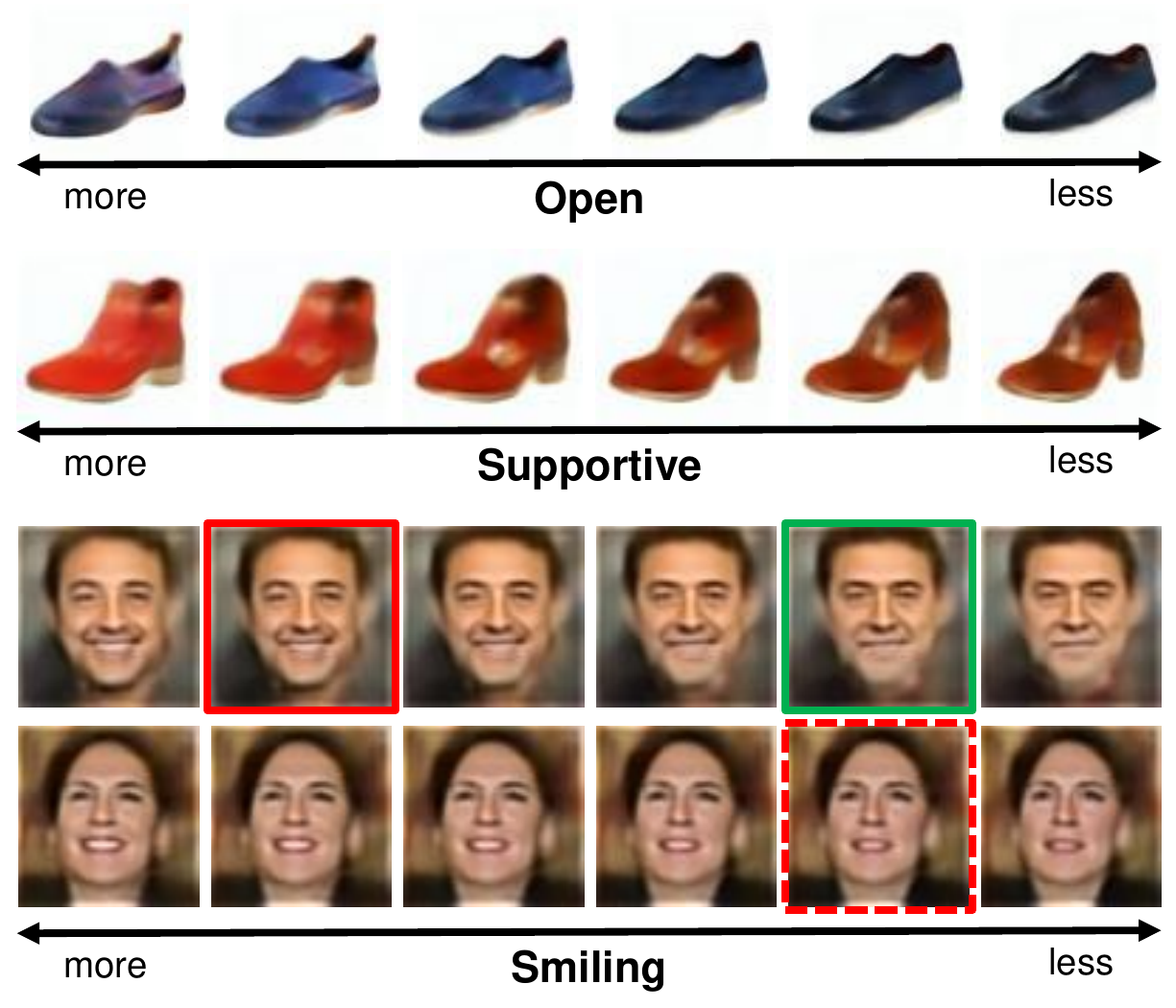}
  \vspace*{-0.05in}
  \caption{Spectra of generated images given an identity and an attribute.  We form two types of image pairs: The two solid boxes represent an \textit{intra-identity pair}, whereas the two red boxes represent an \textit{inter-identity pair.}}
  \vspace*{0.1in}
  \label{fig:spectrum}
\end{figure}

\subsection{Visual Comparison Predictor}
\label{subsec:attributes}
\vspace*{0.05in}

Let $\bm{x}_i \in \mathbb{R}^{N_x}$ denote an image with $N_x$ pixels and let $\phi(\bm{x}_i) \in \mathbb{R}^D$ denote its $D$-dimensional descriptor (e.g., Gist, color, CNN feature, or simply raw pixels).  Given a target attribute $\mathcal{A}$ and two images $\bm{x}_i$ and $\bm{x}_j$, the goal of visual comparison is to determine which of the images contains ``more'' of the specified attribute.

The supervision paradigm proposed for relative attributes~\cite{relative-attributes}, and now widely adopted in various ranking models for attribute comparisons~\cite{Li12,Sandeep14,Singh16,Souri16,Xiao15,Yang16,aron-cvpr2014,aron-iccv2015}, consists of \emph{ordered pairs} of images.  Specifically, the learning algorithm is provided with ordered pairs $\mathcal{P}_\mathcal{A} = \{(\bm{x}_i,\bm{x}_j)\}$ for which human annotators perceive image $i$ to have the attribute $\mathcal{A}$ more than image $j$.  The idea is to learn a ranking function $R_\mathcal{A}(\phi(\bm{x}))$ that satisfies the specified orderings as well as possible:
\vspace*{-0.03in}
\begin{equation}
\forall (i,j) \in \mathcal{P}_\mathcal{A}: R_\mathcal{A}(\phi(\bm{x_i})) > R_\mathcal{A}(\phi(\bm{x_j})).
\vspace*{0.05in}
\end{equation}
Precisely what defines ``as well as possible'' depends on the specifics of the model, such as a RankNet objective~\cite{Singh16,ranknet} or paired classification objective with wide margin regularization~\cite{relative-attributes,Joachims}.

Given a novel image pair $(\bm{x}_m, \bm{x}_n)$, the ranker compares them to determine which exhibits the attribute more.  If $R_\mathcal{A}(\phi(\bm{x}_m)) > R_\mathcal{A}(\phi(\bm{x}_n))$, then image $m$ exhibits attribute $\mathcal{A}$ more than image $n$, and vice versa.

Our goal is to address the sparsity issue in $\mathcal{P}_\mathcal{A}$ through the addition of synthetic image pairs, such that the training pairs are more representative of subtle differences in $\mathcal{A}$.  Our approach does not interfere with the formulation of the specific comparison model used.  So, improvements in densifying supervision are orthogonal to improvements in the relative attribute prediction model.  To demonstrate this versatility, in experiments we explore two successful learning-to-rank models from the attributes literature: one based on RankSVM and another based on a deep Siamese CNN.

\subsection{Synthesizing Dense Supervision}
\label{subsec:generation}
\vspace*{0.05in}

The key to improving \textit{coverage} in the attribute space is the ability to generate images exhibiting subtle differences---with respect to the given attribute---while keeping the others aspects constant.  In other words, we want to walk semantically in the high-level attribute space.

\vspace*{0.1in}
\subsubsection{Attribute-Conditioned Image Generator}
\vspace*{0.05in}

We adopt an existing state-of-the-art image generation system, Attribute2Image, recently introduced by Yan et al.~\cite{Yan16-eccv,Yan16-arxiv}, which can generate images that exhibit a given set of attributes and latent factors.

Suppose we have a lexicon of $N_a$ attributes, $\{\mathcal{A}_1,\dots,\mathcal{A}_{N_a}\}$.
Let $\bm{y} \in \mathbb{R}^{N_a}$ be a vector containing the strength of each attribute, and let $\bm{z} \in \mathbb{R}^{N_z}$ be the latent variables.  The Attribute2Image approach constructs a generative model for
\vspace*{-0.05in}
\begin{equation}
p_\theta(\bm{x} | \bm{y},\bm{z})
\vspace*{0.03in}
\end{equation}
that produces realistic images $\bm{x} \in \mathbb{R}^{N_x}$ conditioned on $\bm{y}$ and $\bm{z}$.  The authors maximize the variational lower bound of the log-likelihood $\text{log}\ p_\theta(\bm{x}|\bm{y})$ in order to obtain the model parameters $\theta$.  The model is implemented with a Conditional Variational Auto-Encoder (CVAE).  The network architecture generates the entangled hidden representation of the attributes and latent factors with multilayer perceptrons, then generates the image pixels with a coarse-to-fine convolutional decoder.  The authors apply their approach for attribute progression, image completion, and image retrieval.  See~\cite{Yan16-eccv,Yan16-arxiv} for more details.

\vspace*{0.1in}
\subsubsection{Generating Dense Synthetic Image Pairs}
\vspace*{0.05in}

We propose to leverage the Attribute2Image~\cite{Yan16-eccv} engine to supply realistic synthetic training images that ``fill in'' underrepresented regions of image space, which we show helps train a model to infer attribute comparisons.

The next key step is to generate a series of synthetic \emph{identities}, then sample images for those identities that are close by in a desired semantic attribute space.\footnote{Note that here the word identity means an instance for some domain, not necessarily a human identity; in experiments we apply our idea both for human faces as well as fashion images of shoes.}  The resulting images will comprise a set of synthetic image pairs $\mathcal{S}_\mathcal{A}$.  We explore two cases for using the generated pairs: one where their putative ordering is verified by human annotators, and another where the ordering implied by the generation engine is taken as their (noisy) label.  Section~\ref{subsec:rankers} describes how we use the hybrid real and synthetic image pairs to train specific attribute predictors.

Each identity is defined by an entangled set of latent factors and attributes.  Let $p(\bm{y})$ denote a prior over the attribute occurrences in the domain of interest.  We model this prior with a multivariate Gaussian whose mean and covariance are learned from the attribute strengths observed in real training images: $p(\bm{y}) = \mathcal{N}(\mu,\Sigma)$.  This distribution captures the joint interactions between attributes, such that a sample from the prior reflects the co-occurrence behavior of different pairs of attributes (e.g., shoes that are very pointy are often also uncomfortable, faces that have facial hair are often masculine, etc.).\footnote{Note that this prior is nonetheless assumed to be coarse, since a subset of dimensions in $\bm{y}$ consist of the very attributes we wish to learn better via densifying supervision.  For the sake of the prior, the training image attribute strengths originate from the raw decision outputs of a preliminary binary attribute classifier trained on disjoint data labeled for the presence/absence of the attribute (see Experiments and Appendix).}  The prior over latent factors $p(\bm{z})$, captures all non-attribute properties like pose, background, and illumination.  Following~\cite{Yan16-arxiv}, we represent $p(\bm{z})$ with an isotropic multivariate Gaussian.

To sample an identity
\begin{equation}
\mathcal{I}_j = (\bm{y}_j,\bm{z}_j)
\end{equation}
we sample $\bm{y}_j$ and $\bm{z}_j$ from their respective priors.  Then, using an Attribute2Image model trained for the domain of interest, we sample from $p_\theta(\bm{x} | \bm{y}_j,\bm{z}_j)$ to generate an image $\hat{\bm{x}}_j \in \mathbb{R}^{N_x}$ for this identity.  Alternatively, we could sample an identity from a single real image, after inferring its latent variables through the generative model~\cite{Yang16}.  However, doing so requires having access to attribute labels for that image.  More importantly, sampling novel identities from the prior (vs.~an individual image) supports our goal to densify supervision, since we can draw nearby instances that need not have been exactly observed in the real training images.  In experiments, we generate thousands of identities.

Next we modify the strength of a single attribute in $\bm{y}$ while keeping all other variables constant.  This yields two ``tweaked" identities $\mathcal{I}_j^{(-)}$ and $\mathcal{I}_j^{(+)}$ that look much like $\mathcal{I}_j$, only with a bit less or more of the attribute, respectively. Specifically, let $\sigma_\mathcal{A}$ denote the standard deviation of attribute scores observed in real training images for attribute $\mathcal{A}$.  We revise the attribute vector for identity $\mathcal{I}_j$ by replacing the dimension for attribute $\mathcal{A}$ according to
\begin{eqnarray}
\nonumber \bm{y}^{(-)}_j(\mathcal{A}) &=& \bm{y}_j(\mathcal{A}) - 2\sigma_\mathcal{A} \text{  and} \\
\bm{y}^{(+)}_j(\mathcal{A}) &=& \bm{y}_j(\mathcal{A}) + 2\sigma_\mathcal{A},
\end{eqnarray}
\vspace*{0.06in}
and $\bm{y}^{(-)}_j(a) = \bm{y}^{(+)}_j(a) = \bm{y}_j(a), \forall a \neq \mathcal{A}$.   Finally, we sample from $p_\theta(\bm{x} | \bm{y}^{(-)}_j,\bm{z}_j)$ and $p_\theta(\bm{x} | \bm{y}^{(+)}_j,\bm{z}_j)$ to obtain images $\hat{\bm{x}}^{(-)}_j$ and $\hat{\bm{x}}^{(+)}_j$.  Recall that our identity sampling accounts for inter-attribute co-occurrences.  Slightly altering a \emph{single} attribute recovers plausible but yet-unseen instances.


\begin{figure}[t]
  \captionsetup{font=footnotesize}
  \centering
  \includegraphics[width=0.48\textwidth]{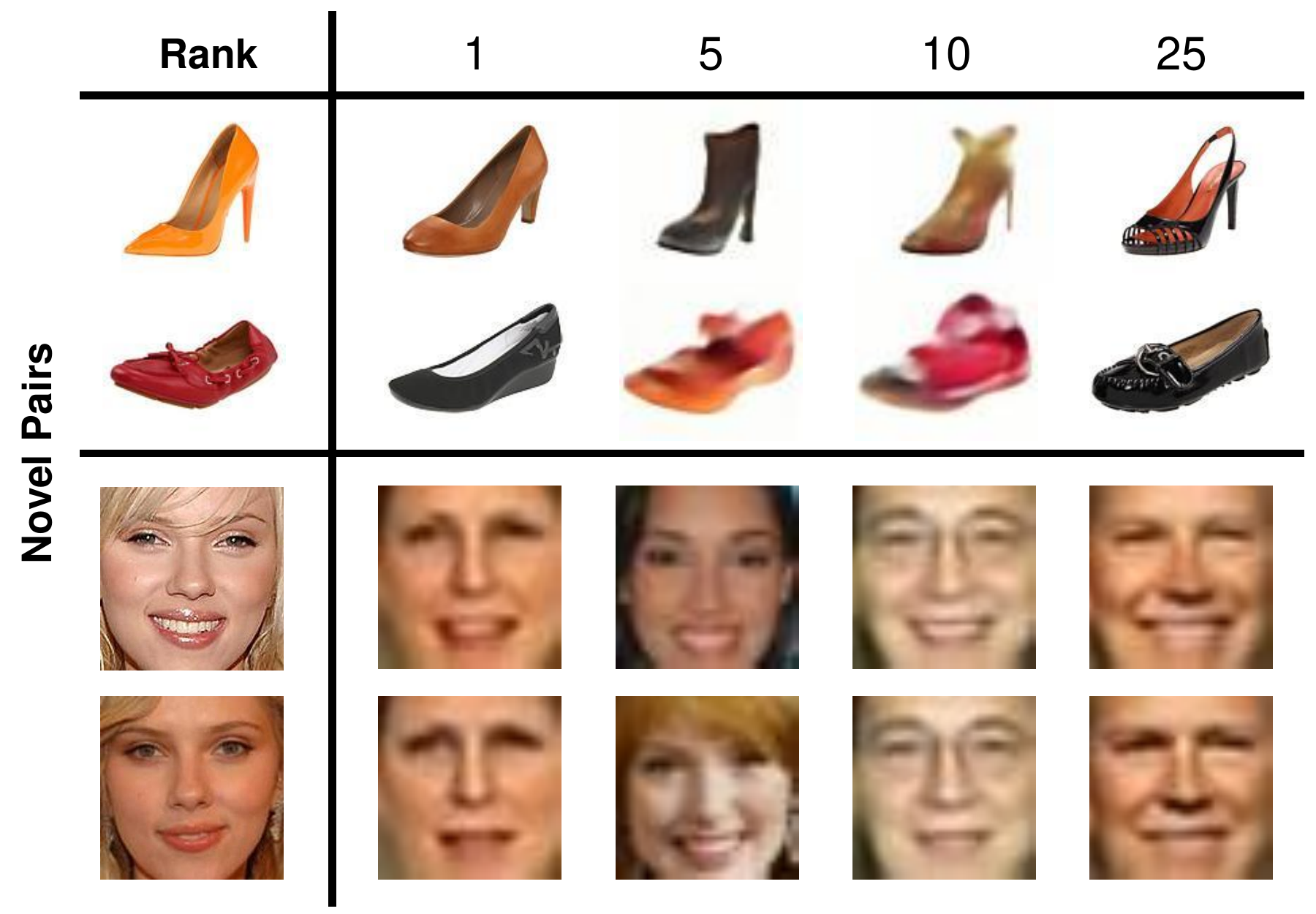}
  \caption{Examples of nearest neighbor image pairs given novel test pairs (left).  Both real and synthetic image pairs appear in the top neighbors, suggesting their combined importance in the local learning algorithm.}
  \vspace*{0.05in}
  \label{fig:qual}
\end{figure}

Figure~\ref{fig:spectrum} shows examples of synthetic images generated for a sampled identity, varying only in one attribute.  The generated images form a smooth progression in the attribute space.  This is exactly what allows us to curate fine-grained pairs of images that are very similar in attribute strength.  Crucially, such pairs are rarely possible to curate systematically among real images.  The exception is special ``hands-on" scenarios, e.g., for faces, asking subjects in a lab to slowly exhibit different facial expressions, or systematically varying lighting or head pose (cf.~PIE, Yale face datasets).  The hands-on protocol is not only expensive, it is inapplicable in most domains outside of faces and for rich attribute vocabularies.  For example, how would one physically modify the pointiness of a shoe's toe, while leaving all other properties the same?  Furthermore, the generation process allows us to collect in a controlled manner subtle visual changes \emph{across} identities as well.

Next we pair up the synthetic images to form the set $\mathcal{S}_\mathcal{A}$, which, once (optionally) verified and pruned by human annotators, will augment the real training image pairs $\mathcal{P}_\mathcal{A}$.\footnote{We also explored training the rankers solely with the synthetic pairs, but find results much weaker than augmenting real pairs with synthetic pairs.  This is likely because the real image pairs play an important role in resisting domain shift problems between the real and synthetic image idiosyncrasies.}  In order to maximize our coverage of the attribute space, we sample two types of synthetic image pairs: \textit{intra-identity pairs}, which are images sampled from the same identity's spectrum and  \textit{inter-identity pairs}, which are images sampled from different spectrums (see Fig.~\ref{fig:spectrum}).  Specifically, for every identity $j$, $\mathcal{S}_\mathcal{A}$ receives intra pairs $\{(\hat{\bm{x}}^{(-)}_j,\hat{\bm{x}}_j), (\hat{\bm{x}}_j,\hat{\bm{x}}^{(+)}_j)\}$ and for every pair of identities $(j,k)$, $\mathcal{S}_\mathcal{A}$ receives inter pairs $\{(\hat{\bm{x}}_j,\hat{\bm{x}}^{(+)}_k), (\hat{\bm{x}}^{(-)}_k,\hat{\bm{x}}_j)\}$.

We expect many of the generated pairs to be valid, meaning that both images are realistic and that the pair exhibits a slight difference in the attribute of interest.  However, this need not always be true.  In some cases the generator will create images that do not appear to manipulate the attribute of interest, or where the pair is close enough in the attribute to be indistinguishable, or where the images simply do not look realistic enough to tell.  Our experiments indicate this happens about 15\% of the time.

To correct erroneous pairs, we collect order labels from 5 crowdworkers per pair.  However, while human-verified pairs are most trustworthy for a learning algorithm, we suspect that even noisy (unverified) pairs could be beneficial too, provided the learning algorithm 1) has high enough capacity to accept a lot of them and/or 2) is label-noise resistant.  Unverified pairs are attractive because they are free to generate in mass quantities.  We examine both cases below.

\vspace*{-0.1in}
\subsection{Learning to Rank with Hybrid Comparisons}
\label{subsec:rankers}
\vspace*{0.05in}

In principle any learning algorithm for visual comparisons could exploit the newly generated synthetic image pairs.  We consider two common ones from the attributes literature: RankSVMs with local learning and a deep Siamese RankNet with a spatial transformer network (STN).

\vspace*{0.1in}
\noindent
\textbf{RankSVM+Local Learning:}  RankSVM is a learning-to-rank solution that optimizes $R_\mathcal{A}(\phi(\bm{x}))$ to preserve orderings of training pairs while maximizing the rank margin, subject to slack~\cite{Joachims}.  In the linear case,
\begin{equation}
R^{(svm)}_\mathcal{A}(\phi(\bm{x})) =\bm{w}_\mathcal{A}^T \phi(\bm{x}),
\label{eq:relattr}
\end{equation}
where $\bm{w}$ is the ranking model parameters.  The formulation is kernelizable, which allows non-linear ranking functions.  It is widely used for attributes~\cite{whittle-ijcv,relative-attributes,aron-cvpr2014,aron-iccv2015,biswas-cvpr2013}.

We employ RankSVM with a \emph{local learning} model.  In local learning, one trains with only those labeled instances  nearest to the test input~\cite{atkeson-1997,bottou-1992}.
Given a hybrid set of sparse real pairs and dense synthetic pairs, $\{\mathcal{P}_\mathcal{A} \bigcup \mathcal{S}_\mathcal{A}\}$, we use a local model to select the most relevant \emph{mix} of real and synthetic pairs (see Fig.~\ref{fig:qual}).  Just as bare bones nearest neighbors relies on adequate density of labeled exemplars to succeed, in general local learning is expected to flourish when the space of training examples is more densely populated.  Thus, local learning is congruent with our hypothesis that data \emph{density} is at least as important as data \emph{quantity} for learning subtle visual comparisons.  See Figure~\ref{fig:concept}.

Specifically, following~\cite{aron-cvpr2014}, we train a local model for each novel image pair (at test time) using only the most relevant image pairs.  Relevance is captured by the inter-pair image distance: for a test pair $(\bm{x}_m,\bm{x}_n)$, one gathers the $K$ nearest pairs according to the product of element-wise distances between $(\bm{x}_m,\bm{x}_n)$ and each training pair.  Only those $K$ pairs are used to train a ranking function (Eqn\eqref{eq:relattr}) to predict the order of $(\bm{x}_m,\bm{x}_n)$.  See~\cite{aron-cvpr2014} for details.

\vspace*{0.1in}
\noindent
\textbf{DeepCNN+Spatial Transformer:}  Our choice for the second ranker is motivated both by its leading empirical performance~\cite{Singh16} as well as its high capacity, which makes it data hungry.

This deep learning to rank method combines a CNN optimized for a paired ranking loss~\cite{ranknet} together with a spatial transformer network (STN)~\cite{STN}.  In particular,
\begin{equation}
R^{(cnn)}_\mathcal{A}(\phi(\bm{x})) = \textrm{RankNet}_\mathcal{A}(\textrm{STN}(\phi(\bm{x}))),
\label{eq:siamese}
\end{equation}
where $\textrm{RankNet}$ denotes a Siamese network with duplicate stacks.  During training these stacks process ordered pairs, learning filters that map the images to scalars that preserve the desired orderings in $\mathcal{P}_\mathcal{A}$.  The STN is trained simultaneously to discover the localized patch per image that is most useful for ranking the given attribute (e.g., it may focus on the mouth for \emph{smiling}).  Given a single novel image, either stack can be used to assign a ranking score.  See~\cite{Singh16} for details.  As above, our approach trains this CNN with all pairs in $\{\mathcal{P}_\mathcal{A} \bigcup \mathcal{S}_\mathcal{A}\}$.

Our approach operates independently of the specific ranking algorithm used for attribute prediction, hence our consideration of two popular methods from the literature.  As we will see below, our results indicate that our idea benefits both.  Furthermore, it has even stronger effects for the higher-capacity deep ranking models that can adequately leverage more densely populated training data.

\vspace*{0.1in}
\noindent
\textbf{Generator vs. Ranker:}  A natural question to ask is why not feed back the synthetic image pairs into the same generative model that produced them, to try and enhance its training?  We avoid doing so for two important reasons.  First, this would lead to a circularity bias where the system would essentially be trying to exploit new data that it has already learned to capture well (and hence could generate already).  Second, the particular image generator we employ is not equipped to learn from relative supervision nor  make relative \emph{comparisons} on novel data.  Rather, it learns from individual images with absolute attribute strengths.  Thus, we use the synthetic data to train a distinct model capable of learning relative visual concepts.

\vspace*{0.1in}
\noindent
\textbf{Curating Images vs. Curating Supervision:}  As discussed in the Introduction, traditional data collection methods lack a direct way to curate image pairs covering the full space of attribute variations, especially those that are fine-grained.  The novelty and strength of our approach lie precisely in addressing this sparsity.  Our approach densifies the attribute space via synthetic images that are plausible photos venturing into potentially undersampled regions of the attribute spectra.  Our approach does not expect to get “something for nothing”.  In particular, it is important to the method design that the synthesized examples will still be processed by human annotators.  Thus, the idea is to expose the learner to realistic images that are critical for fine-grained visual learning yet are difficult to attain in traditional data collection pipelines.

\vspace*{0.1in}
\section{A Lexicon of Fine-Grained Attributes}
\label{sec:lexical}
\vspace*{0.05in}


\begin{figure}[t]
  \captionsetup{font=footnotesize}
  \centering
  \includegraphics[width=0.40\textwidth]{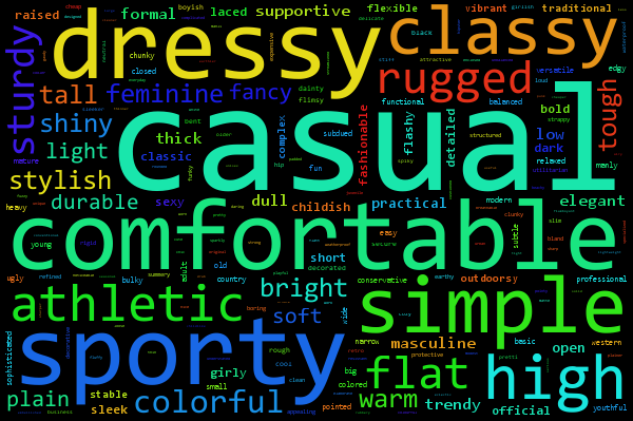}
  \caption{Word cloud depicting our crowd-mined data for a fine-grained relative attribute lexicon for shoes (before post-processing).}
  \label{fig:wordcloud}
\end{figure}

As a secondary contribution, we construct a new fine-grained relative attribute dataset.
As discussed in Sec.~\ref{sec:intro}, label sparsity is an issue for visual comparison.  While there are numerous large  datasets for single image tasks like object detection, datasets for visual comparison with \textit{instance-level pairwise supervision} are more modest.  In addition, the \emph{lexicon} of attributes used in existing relative attributes datasets is selected based on intuitions, i.e., words that seem domain relevant~\cite{whittle-ijcv} or words that seem to exhibit the most subtle fine-grained differences~\cite{aron-cvpr2014}.

Towards addressing both limitations, we 1) use crowdsourcing to mine for an attribute lexicon that is explicitly fine-grained, and 2) collect a large number of pairwise orderings for each attribute in that lexicon.  We focus on fashion images of shoes from the UT-Zap50K dataset~\cite{aron-cvpr2014}.

Given a pair of images, we ask Turkers to complete the sentence, ``Shoe A is a \textit{little more} $\langle$insert word$\rangle$ than Shoe B'' using a single word.  They are instructed to identify subtle differences between the images and provide a short rationale.  The goal is to find out how people differentiate fine-grained differences between shoe images.  Over 1,000 workers participated in the study, yielding a total of 350+ distinct word suggestions across  4,000 image pairs viewed.  This approach to lexicon generation takes inspiration from~\cite{maji-lexicon}, but fine-tuned towards eliciting ``almost indistinguishable" visual changes rather than arbitrary attribute differences.

Figure~\ref{fig:wordcloud} shows a word cloud of the raw results, which we post-process based on the rationales and merging of synonyms. We select the 10 most frequent words as the new \textbf{fine-grained relative attribute lexicon} for shoes:  \textit{comfort}, \textit{casual}, \textit{simple}, \textit{sporty}, \textit{colorful}, \textit{durable}, \textit{supportive}, \textit{bold}, \textit{sleek}, and \textit{open}.  See Appendix for more details.

Using this new lexicon, we collect pairwise supervision for about 4,000 pairs for each of the 10 attributes, using images from UT-Zap50K~\cite{aron-cvpr2014}.  This is a step towards denser supervision on real images---more than three times the comparison labels provided in the original dataset.  Still, as we will see in results, the greater density offered by synthetic training instances is needed for best results.

\vspace*{0.1in}
\section{Experiments}
\label{sec:experiments}
\vspace*{0.05in}

We conduct fine-grained visual comparison experiments to validate the benefit of our dense supervision idea, for both rankers described in Approach.


\begin{table*}[t]
  \captionsetup{font=footnotesize}
  \begin{center}\footnotesize
  \renewcommand{\arraystretch}{1.2}
  \begin{tabular}{|c||l||*{10}{c}|}
  \cline{2-12}
  \multicolumn{1}{c|}{} & & \Gape[4pt]{\textbf{Comfort}} & \textbf{Casual} & \textbf{Simple} & \textbf{Sporty} & \textbf{Colorful} & \textbf{Durable} & \textbf{Supportive} & \textbf{Bold} & \textbf{Sleek} & \textbf{Open} \\
  \Xcline{2-12}{1pt}
  \multicolumn{1}{c|}{} & \Gape[3.2pt]{Classifier} & 72.69 & 79.32 & 82.20 & 81.21 & 17.87 & 80.05 & 78.62 & 18.35 & 17.60 & 27.15 \\
  \hline
  \multirow{4}{*}[-2ex]{\rotatebox{90}{\textbf{RankSVM}}} & Real & 84.03 & 86.11 & 86.89 & 87.27 & 83.84 & 85.15 & \textbf{87.75} & 83.71 & 86.06 & \textbf{84.41} \\
  & Real+ & 82.41 & 87.04 & 86.18 & 87.58 & 84.79 & 84.69 & \textbf{87.75} & 81.44 & 88.02 & 81.18 \\
  & Jitter & 84.49 & 87.35 & 88.52 & 83.36 & 85.36 & 86.77 & 86.86 & 85.36 & 86.31 & 82.53 \\
  & DSynth & \textbf{85.02} & \textbf{88.89} & 85.56 & \textbf{89.95} & \textbf{87.43} & 84.32 & 87.29 & \textbf{87.62} & 86.40 & 81.05 \\
  & DSynth-Auto & 84.72 & 87.35 & \textbf{87.59} & 86.06 & 85.74 & \textbf{86.78} & 83.74 & 85.36 & \textbf{86.55} & 83.87 \\
  \hline
  \multirow{4}{*}[-2ex]{\rotatebox{90}{\textbf{DeepSTN}}} & Real & 84.95 & 87.04 & 89.46 & 88.79 & 94.30 & 83.29 & 85.75 & 87.42 & 85.82 & 84.68 \\
  & Real+ & 81.25 & 87.65 & 86.18 & 87.88 & 90.68 & 83.29 & 85.52 & 87.84 & 86.31 & 82.53 \\
  & Jitter & 81.94 & 87.96 & 86.89 & 87.58 & 93.73 & 85.38 & 85.75 & 89.07 & 83.86 & 80.65 \\
  & DSynth & 82.18 & \textbf{89.81} & \textbf{89.70} & \textbf{90.30} & 93.73 & 87.24 & 85.52 & \textbf{89.28} & 86.55 & 82.26 \\
  & DSynth-Auto & \textbf{87.27} & 88.89 & 88.76 & 90.00 & \textbf{95.44} & \textbf{88.86} & \textbf{87.75} & 87.63 & \textbf{86.80} & \textbf{86.29} \\
  \hline
  \end{tabular}
  \renewcommand{\arraystretch}{1}
  \end{center}
  \vspace*{-0.08in}
  \caption{Results on Zap50K for the new lexicon of 10 attributes most frequently used to distinguish fine-grained differences between shoe images. We experiment with two kinds of base training models: (top) local learning (RankSVM)~\cite{aron-cvpr2014} and (bottom) localized ranking (DeepSTN)~\cite{Singh16}.}
  \label{tab:shoe_results}
\end{table*}

\vspace*{0.1in}
\noindent
\textbf{Datasets}  Our experiments rely on the following existing and newly collected datasets.  To our knowledge there exist no other instance-labeled relative attribute datasets.

\vspace*{0.08in}
\noindent
\textbf{\textit{Zap50K+New Lexicon:}}  The UT-Zap50K dataset~\cite{aron-cvpr2014} consists of 50,025 catalog shoe images from Zappos.com.  It contains 2,800 pairwise labels on average for each of 4 attributes: \textit{open}, \textit{pointy}, \textit{sporty}, and \textit{comfort}.  The labels are divided into coarse (UT-Zap50K-1) and fine-grained pairs (UT-Zap50K-2).  We augment it with the crowd-mined lexicon (cf.~Sec.~\ref{sec:lexical}) for 10 additional attributes.

\vspace*{0.08in}
\noindent
\textbf{\textit{Zap50K-Synth:}}  A new synthetic shoe dataset with pairwise labels on the new 10-attribute lexicon.  We train the generative model using a subset of UT-Zap50K and a superset of the above attributes (see Appendix for details).  We generate 1,000 identities and each one is used to sample both an intra- and inter-identity pair, yielding $\sim$2,000 pair labels per attribute.  The synthetic images are $64\times64$ pixels.

\vspace*{0.08in}
\noindent
\textbf{\textit{LFW-10:}}  The LFW-10 dataset~\cite{Sandeep14} consists of 2,000 face images from  Labeled Faces in the Wild (LFW)~\cite{lfw}.  It contains 10 attributes: \textit{bald}, \textit{dark hair}, \textit{eyes open}, \textit{good looking}, \textit{masculine}, \textit{mouth open}, \textit{smile}, \textit{visible teeth}, \textit{visible forehead}, and \textit{young}.  After pruning pairs with less than 80\% agreement from the workers, there are 600 pairwise labels on average per attribute.

\vspace*{0.08in}
\noindent
\textbf{\textit{PFSmile:}}  Face images from the Public Figures dataset (PubFigAttr)~\cite{relative-attributes,simile}.  8 frontal images each of 8 random individuals are selected, with the frontal images showing different degrees of \textit{smilingness} for the given individual (e.g., images of Zach Efron going from not smiling at all to fully smiling).  We use smiling because it is the only PubFig attribute that manifests fine-grained changes on the same individual (e.g., it doesn't display Zach both as \textit{bald} and \textit{less bald}). This limitation of the data actually reinforces the difficulty of manually curating images with subtle differences for learning.  We collect labels on all possible pairwise comparisons among images of the same individual.  After pruning, there are 211 pairwise labels.

\vspace*{0.08in}
\noindent
\textbf{\textit{LFW-Synth:}}  A new synthetic face dataset with pairwise labels on the attribute \textit{smiling}.  We train the generative model on a subset of LFW images and  the 73 attributes from~\cite{Yan16-eccv,simile}.  We generate 2,000 identities and sample a total of 4,000 intra pairs and 1,000 inter pairs.  The synthetic images are $35\times35$ pixels, after zooming to a tight bounding box around the face region.

\vspace*{0.1in}
\noindent
\textbf{Implementation Details}  We downsize real images to match the resolution of the synthetic ones.  We use the code kindly shared by the authors for the Attribute2Image system~\cite{Yan16-eccv}, with all default parameters including the prior on $\bm{z}$.  Early experiments showed that a mix of inter and intra-identity pairs was most effective, so we use a 50-50 mix in all experiments.  For  RankSVM, we use Gist~\cite{gist} and 30-bin Lab color histograms as the image features $\phi$, following~\cite{relative-attributes,aron-cvpr2014}\footnote{Pretrained CNN features with RankSVM proved inferior.}, and validate $K$ per method on held-out data.  For DeepSTN, we use training parameters provided in~\cite{Singh16} per dataset, and $\phi$ is simply the pixels.  The images used to train the generative model, to train the ranking functions, and to evaluate (test set) are kept strictly disjoint.

\vspace*{0.13in}
\noindent
\textbf{Baselines}  We compare the following methods:
\vspace*{0.05in}

\begin{itemize}

  \item \textbf{Real}:  Training pool consists of only real image pairs, labeled by human annotators.
  \vspace*{0.1in}
  \item \textbf{Jitter}:  Uses the same real training pairs, but augments them with pairs using traditional low-level jitter.  Each real image is jittered following parameters in~\cite{Dosovitskiy14} in a combination of five changes: translation, scaling, rotation, contrast, and color.  A jittered pair inherits the corresponding real pair's label.
  \vspace*{0.1in}
  \item \textbf{DSynth}:  Training pool consists of only \emph{half} of Real's pairs, with the other half replaced with our dense synthetic image pairs, manually verified by annotators.
  \vspace*{0.1in}
  \item \textbf{DSynth-Auto}:  Training pool consists of all real image pairs and our automatically supervised synthetic image pairs, where \textit{noisy} pairwise supervision is obtained (for free) based on the absolute attribute strength used to generate the respective images.  We explore this variant of our approach to see in practice how essential human supervision is for the synthetic image pairs.  The role of our method is to densify the training data, not to get labels for free.  Nonetheless, we speculate that damage done by noisy labels may in some cases be favorably balanced out by the better coverage of relevant data samples.\footnote{We find that the auto labels have a 78\% agreement on average with the human labels across all attributes.}
  \vspace*{0.1in}
  \item \textbf{Classifier}:  Predicts the attribute scores directly using the posterior $R_\mathcal{A}(\phi(\bm{x})) = p(\mathcal{A}|\bm{x})$ obtained from a binary classifier trained with the same images that train the image generator.
  \vspace*{0.1in}
  \item \textbf{Real+}:  Augments Real with additional pseudo real image pairs.  Recall that the image generator requires attribute strength values on its training images, which are obtained from outputs of an attribute classifier~\cite{simile}.  The Real+ baseline trains its ranking function using the same real image pairs used above, plus pseudo pairs of the equal size boostrapped from those strength values on individual images.
  \vspace*{0.1in}
\end{itemize}

\noindent
We stress that our DSynth methods use the same amount of labeled ordered pairs as the Real and Jitter baselines.

\subsection{Fashion Images of Shoes}
\vspace*{0.05in}

Fashion product images offer a great testbed for fine-grained comparisons.  This experiment uses UT-Zap50K for real training and testing pairs, and UT-Zap50K-Synth for synthetic training pairs.  There are 10 attributes total.  Since the real train and test pairs come from the same dataset, this presents a challenge for our approach---can synthetic images, despite their inherent domain shift, still help the algorithm learn a more reliable model?

Table~\ref{tab:shoe_results} shows the results.  The Classifier baseline underperforms both rankers, confirming that the generator's initial representation of attribute strengths is insufficient.

Under the local RankSVM model, our approach outperforms the baselines in most attributes.  Augmenting with traditional low-level jitter also provides a slight boost in performance, but not as much as ours.  Looking at the composition of the local neighbors, we see that about 85\% of the selected local neighbors are our synthetic pairs (15\% real) while only 55\% are jittered pairs (45\% real).  Thus, our synthetic pairs do indeed heavily influence the learning of the ranking models.  Figure~\ref{fig:qual} shows examples of nearest neighbor image pairs retrieved for sample test pairs.  The examples illustrate how 1) the synthetic images densify the supervision, providing perceptually closer instances for training, and 2) that both real and synthetic image pairs play an important role in the local learning algorithm.  We conclude that semantic jitter densifies the space more effectively than low-level jitter.

Under the DeepSTN model (Table~\ref{tab:shoe_results}), our approach outperforms the baselines in all attributes, demonstrating the strength of our dense synthetic pairs in a high capacity model.  Interestingly, DSynth-Auto often outperforms DSynth here.  We believe the higher capacity of the DeepSTN model can better leverage the noisy auto-labeled pairs, compared to the RankSVM model, which more often benefits from the human-verification step.  As one would expect, we notice that DSynth-Auto does best for attributes where the inferred labels agree most often with human provided labels. This is an exciting outcome; our model has potential to generate useful training data with ``free'' supervision.  Low-level jitter on the other hand has limited benefit, even detrimental in some cases.

Additionally, we experiment with an extra baseline Real+.  We convert the generator's initial representation into pairwise labels by looking at the decision values (i.e. attribute strength) from the trained classifiers.  We combine these \textit{pseudo} pairs with real pairs for training.  As we see in Table~\ref{tab:shoe_results}, the additional information does not provide any performance gain over the original Real baseline.  We believe Real+ suffers from the same sparsity issue as Real and is not providing the fine-grained comparisons needed to train a stronger model.

Overall, our gains are significant, considering they are achieved without any changes to the underlying ranking models, the features, or the experimental setup.


\begin{table*}[t]
  \captionsetup{font=footnotesize}
  \begin{center}\footnotesize
  \renewcommand{\arraystretch}{1.2}
  \begin{tabular}{|l||*{10}{c}||c|}
  \hline
   & \Gape[4pt]{\textbf{Comfort}} & \textbf{Casual} & \textbf{Simple} & \textbf{Sporty} & \textbf{Colorful} & \textbf{Durable} & \textbf{Supportive} & \textbf{Bold} & \textbf{Sleek} & \textbf{Open} & \textbf{Smile} \\
  \Xhline{1pt}
  Real & 85.09 & 74.48 & 83.59 & 83.93 & 99.31 & 91.16 & 81.03 & 87.59 & 78.76 & 85.83 & 87.67 \\
  DSynth & \textbf{95.96} & \textbf{89.58} & \textbf{90.77} & \textbf{91.96} & \textbf{99.77} & \textbf{99.77} & \textbf{90.12} & \textbf{93.98} & \textbf{98.12} & \textbf{91.09} & \textbf{96.67} \\
  DSynth-Auto & 93.17 & 78.65 & \textbf{90.77} & 73.66 & 99.54 & \textbf{99.77} & 79.84 & 89.47 & \textbf{98.12} & 79.35 & 96.00 \\
  \hline
  \end{tabular}
  \renewcommand{\arraystretch}{1}
  \end{center}
  \vspace*{-0.08in}
  \caption{Results on human-labeled synthetic test pairs for both domains using the DeepSTN model.}
  \label{tab:synthtest_results}
\end{table*}


\begin{table}[t]
  \captionsetup{font=footnotesize}
  \begin{center}\footnotesize
  \renewcommand{\arraystretch}{1.2}
  \begin{tabular}{|c||l||c c c|}
  \cline{2-5}
  \multicolumn{1}{c|}{} & & \Gape[4pt]{\textbf{Open}} & \textbf{Sporty} & \textbf{Comfort} \\
  \Xhline{1pt}
  \multirow{4}{*}{\rotatebox[origin=c]{90}{\textbf{Zap50K-1}}} & RelAttr~\cite{relative-attributes} & 88.33 & 89.33 & 91.33 \\
  & FG-LP~\cite{aron-cvpr2014} & 90.67 & 91.33 & 93.67 \\
  & DeepSTN~\cite{Singh16} & 93.00 & 93.67 & 94.33 \\
  \cline{2-5}
  & DSynth-Auto (Ours) & \textbf{95.00} & \textbf{96.33} & \textbf{95.00} \\
  \Xhline{1pt}
  \multirow{4}{*}{\rotatebox[origin=c]{90}{\textbf{Zap50K-2}}} & RelAttr~\cite{relative-attributes} & 60.36 & 65.65 & 62.82 \\
  & FG-LP~\cite{aron-cvpr2014} & 69.36 & 66.39 & 63.84 \\
  & DeepSTN~\cite{Singh16} & 70.73 & 67.49 & 66.09 \\
  \cline{2-5}
  & DSynth-Auto (Ours) & \textbf{72.18} & \textbf{68.70} & \textbf{67.72} \\
  \hline
  \end{tabular}
  \renewcommand{\arraystretch}{1}
  \end{center}
  \vspace*{-0.08in}
  \caption{Results on UT-Zap50K-1 (coarse pairs) and UT-Zap50K-2 (fine-grained pairs) vs. prior methods.  Note that all methods are trained and tested on $64\times 64$ images for an apples-to-apples comparison.  All experimental setup are kept the same except for the addition of dense synthetic pairs to the training pool for our approach.}
  \vspace*{0.15in}
  \label{tab:zap50k_rerun}
\end{table}


\begin{table}[t]
  \captionsetup{font=footnotesize}
  \begin{center}\footnotesize
  \renewcommand{\arraystretch}{1.2}
  \begin{tabular}{|l||c c c c c|}
  \hline
  \textbf{Smiling} & Real & Real+ & Jitter & DSynth & DSynth-Auto \\
  \Xhline{1pt}
  Classifier & \multicolumn{5}{ c| }{ --\quad--\quad--\quad--\quad--\quad--\quad62.35\quad--\quad--\quad--\quad--\quad--\quad-- }  \\
  \hline
  RankSVM & 69.29 & 68.95 & 74.29 & 73.88 & \textbf{75.00} \\
  \hline
  DeepSTN & 81.52 & 80.84 & 80.09 & \textbf{85.78} & 84.36 \\
  \hline
  \end{tabular}
  \renewcommand{\arraystretch}{1}
  \end{center}
  \vspace*{-0.08in}
  \caption{Results on PFSmile dataset.}
  \vspace*{0.1in}
  \label{tab:face_results}
\end{table}

\noindent
\textbf{Comparison to prior relative attribute results}  Next, we take the best model from above (DeepSTN+DSynth-Auto), and compare its results to several existing methods.  While authors have reported accuracies on this dataset, as-is comparisons to our model would not be apples-to-apples: due to the current limits of image synthesis, we work with low resolution data ($64\times 64$) whereas prior work uses full sized $150 \times 100$ ~\cite{Singh16,Souri16,Xiao15}.  Therefore, we use available authors' code to re-train existing methods from scratch with the same smaller real images we use.  We zero-pad the real images to squares to preserve the aspect ratios.  In particular, we train 1) Relative attributes (\textbf{RelAttr})~\cite{relative-attributes}; 2) Fine-grained local learning (\textbf{FG-LP})~\cite{aron-cvpr2014}; and 3) End-to-end localization and ranking (\textbf{DeepSTN}). We compare them on UT-Zap50K, a primary benchmark for relative attributes~\cite{aron-cvpr2014}.\footnote{We test all attributes that overlap between UT-Zap50K and our newly collected lexicon of 10.}  We do \emph{not} use the newly collected real labeled data for our method, to avoid an unfair advantage.

Tables~\ref{tab:zap50k_rerun} shows the results.  Our approach beats all of them, outperforming the state-of-the-art DeepSTN even for the difficult fine-grained pairs on UT-Zap50K-2 where attention to subtle details is necessary.

\subsection{Human Faces}
\vspace*{0.05in}

Next we consider the face domain, where fine-grained comparisons are also of great practical interest.  Since PFSmile only contains image pairs of the same individual, the comparison task is fine-grained by design.  This experiment uses LFW-10 for real training pairs, LFW-Synth for synthetic training pairs, and PFSmile for real testing pairs.  Here we have an additional domain shift, since the real train and test images are from different datasets with somewhat different properties.

Table~\ref{tab:face_results} shows the results.  Consistent with above, our approach outperforms all baselines. Even without human verification of our synthetic pairs (DSynth-Auto), our method secures a decent gain over the Real baseline: 75.00\% vs. 69.29\% and 84.36\% vs. 81.52\%.  That amounts to a relative gain of 8\% and 3.5\%, respectively.  The Classifier posterior baseline underperforms the rankers.  Our semantic jitter strongly outperforms traditional low-level jitter for the DeepSTN rankers, with a 6 point accuracy boost.

\subsection{Synthetic Test Images}
\vspace*{0.05in}

Finally, to see the effect of domain shifts, we consider test sets for both datasets comprised of novel synthetic image pairs drawn from 2,000 identities.  The ordering labels are human verified for all test image pairs.  Table~\ref{tab:synthtest_results} shows the results using the DeepSTN model.  The RankSVM model achieves similar results.

Our gains here are substantial, e.g., an 25 point absolute gain for \emph{sleek}.  Admittedly, our method has the advantage here of learning on data from the same domain (namely, synthetic images generated from Attribute2Image), whereas the Real baseline has to overcome this domain shift to generalize.  However, our method overcomes this very same domain shift in the other direction in all of the results reported above on real test image pairs.

\section{Conclusion}
\label{sec:conclusion}
\vspace*{0.05in}

Supervision sparsity hurts fine-grained attributes---closely related image pairs are exactly the ones the system must learn from.  We presented a new approach to training data augmentation, in which real training data mixes with realistic synthetic examples that vary slightly in their attributes.  The generated training images more densely sample the space of images to illustrate fine-grained differences.  We stress that sample \emph{density} is distinct from sample \emph{quantity}.  As our experiments demonstrate, simply gathering more real images does not offer the same fine-grained density, due to the curation problem.  We are also interested in exploring semantic jitter for other recognition tasks.

Future work could explore ways to minimize the effort to only the most questionable synthetic pairs, in order to augment the real pairs with a mix of automatically generated and human generated comparisons.

\vspace*{0.1in}
{\small
\bibliographystyle{IEEEtran}
\bibliography{strings,refs}

\begin{thebibliography}{10}
\providecommand{\url}[1]{#1}
\csname url@samestyle\endcsname
\providecommand{\newblock}{\relax}
\providecommand{\bibinfo}[2]{#2}
\providecommand{\BIBentrySTDinterwordspacing}{\spaceskip=0pt\relax}
\providecommand{\BIBentryALTinterwordstretchfactor}{4}
\providecommand{\BIBentryALTinterwordspacing}{\spaceskip=\fontdimen2\font plus
\BIBentryALTinterwordstretchfactor\fontdimen3\font minus
  \fontdimen4\font\relax}
\providecommand{\BIBforeignlanguage}[2]{{%
\expandafter\ifx\csname l@#1\endcsname\relax
\typeout{** WARNING: IEEEtran.bst: No hyphenation pattern has been}%
\typeout{** loaded for the language `#1'. Using the pattern for}%
\typeout{** the default language instead.}%
\else
\language=\csname l@#1\endcsname
\fi
#2}}
\providecommand{\BIBdecl}{\relax}
\BIBdecl

\bibitem{Datta11}
A.~Datta, R.~Feris, and D.~Vaquero, ``Hierarchical ranking of facial
  attributes,'' in \emph{Face and Gesture}, 2011.

\bibitem{whittle-ijcv}
A.~Kovashka, D.~Parikh, and K.~Grauman, ``Whittle{S}earch: Interactive image
  search with relative attribute feedback,'' \emph{International Journal of
  Computer Vision (IJCV)}, vol. 115, no.~2, pp. 185--210, Nov 2015.

\bibitem{Li12}
S.~Li, S.~Shan, and X.~Chen, ``Relative forest for attribute prediction,'' in
  \emph{ACCV}, 2012.

\bibitem{relative-attributes}
D.~Parikh and K.~Grauman, ``Relative {A}ttributes,'' in \emph{Proceedings of
  the IEEE International Conference on Computer Vision (ICCV)}, 2011.

\bibitem{Sandeep14}
R.~Sandeep, Y.~Verma, and C.~Jawahar, ``Relative parts: Distinctive parts for
  learning relative attributes,'' in \emph{CVPR}, 2014.

\bibitem{gupta-eccv2012}
A.~Shrivastava, S.~Singh, and A.~Gupta, ``Constrained semi-supervised learning
  using attributes and comparative attributes,'' in \emph{ECCV}, 2012.

\bibitem{Singh16}
K.~Singh and Y.~J. Lee, ``End-to-end localization and ranking for relative
  attributes,'' in \emph{ECCV}, 2016.

\bibitem{Souri16}
Y.~Souri, E.~Noury, and E.~Adeli, ``Deep relative attributes,'' in \emph{ACCV},
  2016.

\bibitem{Xiao15}
F.~Xiao and Y.~J. Lee, ``Discovering the spatial extent of relative
  attributes,'' in \emph{ICCV}, 2015.

\bibitem{Yang16}
X.~Yang, T.~Zhang, C.~Xu, S.~Yan, M.~Hossain, and A.~Ghoneim, ``Deep relative
  attributes,'' \emph{IEEE Trans. on Multimedia}, vol.~18, no.~9, Sept 2016.

\bibitem{aron-cvpr2014}
A.~Yu and K.~Grauman, ``Fine-grained visual comparisons with local learning,''
  in \emph{Proceedings of the IEEE Conference on Computer Vision and Pattern
  Recognition (CVPR)}, 2014.

\bibitem{aron-iccv2015}
------, ``Just noticeable differences in visual attributes,'' in
  \emph{Proceedings of the IEEE International Conference on Computer Vision
  (ICCV)}, 2015.

\bibitem{imagenet}
J.~Deng, W.~Dong, R.~Socher, L.-J. Li, K.~Li, and L.~Fei-Fei, ``Imagenet: A
  {L}arge-{S}cale {H}ierarchical {I}mage {D}atabase,'' in \emph{Proceedings of
  the IEEE Conference on Computer Vision and Pattern Recognition (CVPR)}, 2009.

\bibitem{biswas-cvpr2013}
A.~Biswas and D.~Parikh, ``Simultaneous active learning of classifiers and
  attributes via relative feedback,'' in \emph{Proceedings of the IEEE
  Conference on Computer Vision and Pattern Recognition (CVPR)}, 2013.

\bibitem{Dosovitskiy14}
A.~Dosovitskiy, J.~Springenberg, M.~Riedmiller, and T.~Brox, ``Discriminative
  unsupervised feature learning with convolutional neural networks,'' in
  \emph{NIPS}, 2014.

\bibitem{Simard03}
P.~Simard, D.~Steinkraus, and J.~Platt, ``Best practices for convolutional
  neural networks applied to visual document analysis,'' in \emph{ICDAR}, 2003.

\bibitem{Vincent08}
P.~Vincent, H.~Larochelle, Y.~Bengio, and P.~Manzagol, ``Extracting and
  composing robust features with denoising autoencoders,'' in \emph{ICML},
  2008.

\bibitem{nixon-attributes}
D.~Reid and M.~Nixon, ``Human identification using facial comparative
  descriptions,'' in \emph{ICB}, 2013.

\bibitem{adobe-relative-attributes-fonts}
P.~O'Donovan, J.~Libeks, A.~Agarwala, and A.~Hertzmann, ``Exploratory font
  selection using crowdsourced attributes,'' in \emph{SIGGRAPH}, 2014.

\bibitem{Joachims}
T.~Joachims, ``Optimizing search engines using clickthrough data,'' in
  \emph{SIGKDD}, 2002.

\bibitem{Khosla13}
A.~Khosla, W.~A. Bainbridge, A.~Torralba, and A.~Oliva, ``Modifying the
  memorability of face photographs,'' in \emph{International Conference on
  Computer Vision (ICCV)}, 2013.

\bibitem{Kemelmacher14}
I.~Kemelmacher-Shlizerman, S.~Suwajanakorn, and S.~Seitz, ``Illumination-aware
  age progression,'' in \emph{CVPR}, 2014.

\bibitem{Laffont14}
P.~Laffont, Z.~Ren, X.~Tao, C.~Qian, and J.~Hays, ``Transient attributes for
  high-level understanding and editing of outdoor scenes,'' in \emph{SIGGRAPH},
  2014.

\bibitem{Goodfellow14}
I.~Goodfellow, J.~Pouget-Abadie, M.~Mirza, B.~Xu, D.~Warde-Farley, S.~Ozair,
  A.~Courville, and Y.~Bengio, ``Generative adversarial nets,'' in \emph{NIPS},
  2014.

\bibitem{Gregor15}
K.~Gregor, I.~Danihelka, A.~Graves, and D.~Wierstra, ``Draw: A recurrent neural
  network for image generation,'' in \emph{ICML}, 2015.

\bibitem{Kingma14}
D.~Kingma and M.~Welling, ``Auto-encoding variational bayes,'' in \emph{ICLR},
  2014.

\bibitem{Kulkarni15}
T.~Kulkarni, W.~Whitney, P.~Kohli, and J.~Tenenbaum, ``Deep convolutional
  inverse graphics network,'' in \emph{NIPS}, 2015.

\bibitem{Dosovitskiy15}
A.~Dosovitskiy, J.~Springenberg, and T.~Brox, ``Learning to generate chairs
  with convolutional neural networks,'' in \emph{CVPR}, 2015.

\bibitem{Li16}
M.~Li, W.~Zuo, and D.~Zhang, ``Convolutional network for attribute-driven and
  identity-preserving human face generation,'' Tech. Rep. arXiv:1608.06434,
  2016.

\bibitem{Pandey16}
G.~Pandey and A.~Dukkipati, ``Variational methods for conditional multimodal
  learning: Generating human faces from attributes,'' Tech. Rep.
  arXiv:1603.01801, 2016.

\bibitem{Yan16-eccv}
X.~Yan, J.~Yang, K.~Sohn, and H.~Lee, ``Attribute2{I}mage: Conditional image
  generation from visual attributes,'' in \emph{Proceedings of European
  Conference on Computer Vision (ECCV)}, 2016.

\bibitem{Yan16-arxiv}
------, ``Attribute2{I}mage: Conditional image generation from visual
  attributes,'' Tech. Rep. arXiv:1512.00570, 2016.

\bibitem{psh}
G.~Shakhnarovich, P.~Viola, and T.~Darrell, ``Fast {P}ose {E}stimation with
  {P}arameter-{S}ensitive {H}ashing,'' in \emph{Proceedings of the IEEE
  International Conference on Computer Vision (ICCV)}, 2003.

\bibitem{shotton-cvpr2011}
J.~Shotton, A.~Fitzgibbon, M.~Cook, T.~Sharp, M.~Finocchio, R.~Moore,
  A.~Kipman, and A.~Blake, ``Real-time human pose recognition in parts from
  single depth images,'' in \emph{CVPR}, 2011.

\bibitem{pishchulin-bmvc2011}
L.~Pishchulin, A.~Jain, C.~Wojek, T.~Thormahlen, and B.~Schiele, ``In good
  shape: Robust people detection based on appearance and shape,'' in
  \emph{BMVC}, 2011.

\bibitem{ParkRamanan}
D.~Park and D.~Ramanan, ``Articulated pose estimation with tiny synthetic
  videos,'' in \emph{ChaLearn Workshop, CVPR}, 2015.

\bibitem{saenko-iccv2015}
X.~Peng, B.~Sun, K.~Ali, and K.~Saenko, ``Learning deep object detectors from
  3d models,'' in \emph{ICCV}, 2015.

\bibitem{jaderberg-text-2014}
M.~Jaderberg, K.~Simonyan, A.~Vedaldi, and A.~Zisserman, ``Synthetic data and
  artificial neural networks for natural scene text recognition,'' in
  \emph{NIPS 14 Deep Learning workshop}, 2014.

\bibitem{ranknet}
C.~Burges, T.~Shaked, E.~Renshaw, A.~Lazier, M.~Deeds, N.~Hamilton, and
  G.~Hullender, ``Learning to rank using gradient descent,'' in \emph{ICML},
  2005.

\bibitem{atkeson-1997}
C.~Atkeson, A.~Moore, and S.~Schaal, ``Locally weighted learning,''
  \emph{Artificial Intelligence Review}, vol.~11, pp. 11--73, 1997.

\bibitem{bottou-1992}
L.~Bottou and V.~Vapnik, ``Local learning algorithms,'' \emph{Neural
  Computation}, 1992.

\bibitem{STN}
M.~Jaderberg, K.~Simonyan, A.~Zisserman, and K.~Kavukcuoglu, ``Spatial
  transformer networks,'' in \emph{NIPS}, 2015.

\bibitem{maji-lexicon}
S.~Maji, ``Discovering a lexicon of parts and attributes,'' in \emph{Second
  International Workshop on Parts and Attributes, ECCV}, 2012.

\bibitem{lfw}
G.~B. Huang, M.~Ramesh, T.~Berg, and E.~Learned-Miller, ``Labeled faces in the
  wild: A database for studying face recognition in unconstrained
  environments,'' University of Massachusetts, Amherst, Tech. Rep. 07-49,
  October 2007.

\bibitem{simile}
N.~Kumar, A.~Berg, P.~Belhumeur, and S.~Nayar, ``{Attribute and Simile
  Classifiers for Face Verification},'' in \emph{Proceedings of the IEEE
  International Conference on Computer Vision (ICCV)}, 2009.

\bibitem{gist}
A.~Torralba, ``Contextual {P}riming for {O}bject {D}etection,''
  \emph{International Journal of Computer Vision (IJCV)}, vol.~53, no.~2, pp.
  169--191, 2003.

\end{thebibliography}
}

\newpage
\newpage
\newpage
\clearpage

\section{Appendix}
\label{sec:appendix}
\vspace*{0.05in}

\subsection{Fine-Grained Attribute Lexicon}

We use the UT-Zap50K shoe dataset~\cite{aron-cvpr2014} to perform our lexical study.  It contains 50,025 catalog shoe images along with a set of meta-data that are associated with each image.  Our goal is to study how humans distinguish fine-grained differences in similar images.  Specifically, we want to know what words humans use to describe fine-grained differences.

\vspace*{0.1in}
\noindent
\textbf{Experimental Design}  We design our experiments in the form of ``complete the sentence'' questions and test them on the Amazon MTurk workers.  We experiment with two kinds of designs: Design 1 compares two individual images while Design 2 compares one image against a group of six images.  Given the meta-data which contains a category (i.e. slippers, boots) and subcategory (i.e. flats, ankle high) labels for each image, we combine these labels into a set of 21 unique category-subcategory pseudo-classes (i.e. slippers-flats, shoes-loaders).  Using theses new pseudo-classes, we sample 4,000 supervision pairs (for each design) where 80\% are comparing within the same pseudo-class and 20\% are comparing within the same category.  By focusing sampled pairs among items within a pseudo-class, we aim for a majority of the pairs to contain visually quite related items, thus forcing the human subjects to zero in on fine-grained differences.

\begin{figure*}[t]
  \captionsetup{font=footnotesize}
  \centering
  \includegraphics[width=0.75\textwidth]{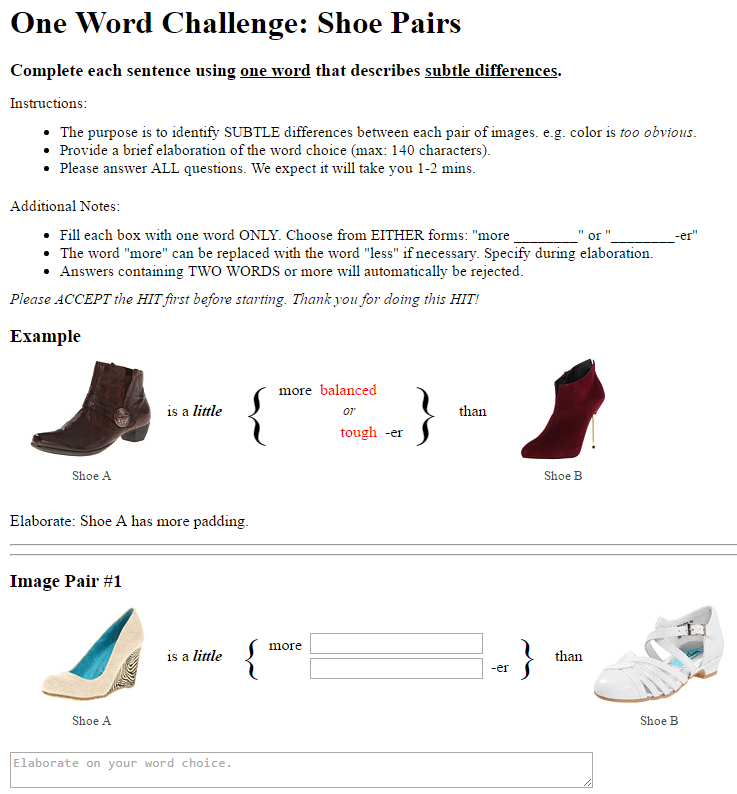}
  \caption{Screenshot of our lexical experiment on MTurk in Design 1.}
  \label{fig:hit}
\end{figure*}

For each question, the workers are asked to complete the sentence, ``Shoe A is a \textit{little more/less} $\langle$insert word$\rangle$ than Shoe B'' using a single word (``Shoe B'' is replaced by ``Group B'' for Design 2).  They are instructed to identify \textbf{subtle differences} between the images and provide a short rationale to elaborate on their choices.  Figure~\ref{fig:hit} shows a screenshot of a sample question.

\vspace*{0.1in}
\noindent
\textbf{Post-Processing}  We post-process the fine-grained word suggestions through correcting for human variations (i.e. misspelling, word forms), merging of visual synonyms/antonyms, and evaluation of the rationales.  For example, ``casual'' and ``formal'' are visual antonyms and workers used similar keywords in their rationales for ``durable'' and ``rugged''.  In both cases, the frequency counts for the two words are combined.  Over 1,000 MTurk workers participated in our study, yielding a total of 350+ distinct word suggestions\footnote{We used only the words from Design 1 as the two designs produced very similar word suggestions.}.  In the end, we select the 10 most frequently appearing words as our fine-grained relative attribute lexicon for shoes:  \textit{comfort}, \textit{casual}, \textit{simple}, \textit{sporty}, \textit{colorful}, \textit{durable}, \textit{supportive}, \textit{bold}, \textit{sleek}, and \textit{open}.

\vspace*{0.05in}
\subsection{Generative Model Training}
\vspace*{0.05in}

We train our attribute-conditioned image generator using a Conditional Variational Auto-Encoder (CVAE)~\cite{Yan16-eccv}.  The model requires a vector of real-valued attribute strengths for each training image.  We detail the setup process for each dataset below.

\noindent
\textbf{Fashion Images of Shoes}  We use a subset of 38,866 images from UT-Zap50K to train the generative shoe model.  Using the meta-data once again, we select 40 attributes ranging from material types to toe styles (e.g. Material.Mesh, ToeStyle.Pointed, etc.) and assign binary pre-labels to them.  In addition, we also use the 10 fine-grained relative attributes collected from our lexical study.  We sample 500 supervision pairs for each attribute from the newly collected pairwise labels and train linear SVM rankers using RankSVM~\cite{Joachims}.  We then project all 1,000 images (used to train the ranker) onto the learned ranker to obtain their real-valued ranking scores, which we use as their pre-labels.  While our focus is on the 10 relative attributes, the inclusion of additional attributes aids in overall learning of the generative model.  However, we do not use any of those meta-data attributes for fine-grained relative attribute training as they are mostly binary in nature.

Finally, using these pre-labels from all 50 attributes, we train a linear classifier for each attribute.  We apply the classifier on all 38,866 images and use their decision values as the real-valued attribute strength needed to train the generative model.  All of this is a workaround, similar to the one used in~\cite{Yan16-eccv}, in order to supply the generative model with real-valued attribute strengths on its training data.  If labeled binary attribute data were available for training the linear classifiers from the onset, that would be equally good if not better.

\vspace*{0.1in}
\noindent
\textbf{Human Faces}  We use a subset of 11,154 images from LFW~\cite{lfw} to train the generative face model.  Following~\cite{Yan16-eccv}, we use the 73 dimensional attribute strength provided in~\cite{simile} to train the generative face model.

\vspace*{0.1in}
\subsection{Nearest Neighbors}
\vspace*{0.05in}

\begin{figure*}[t]
  \captionsetup{font=footnotesize}
  \centering
  \includegraphics[width=0.7\textwidth]{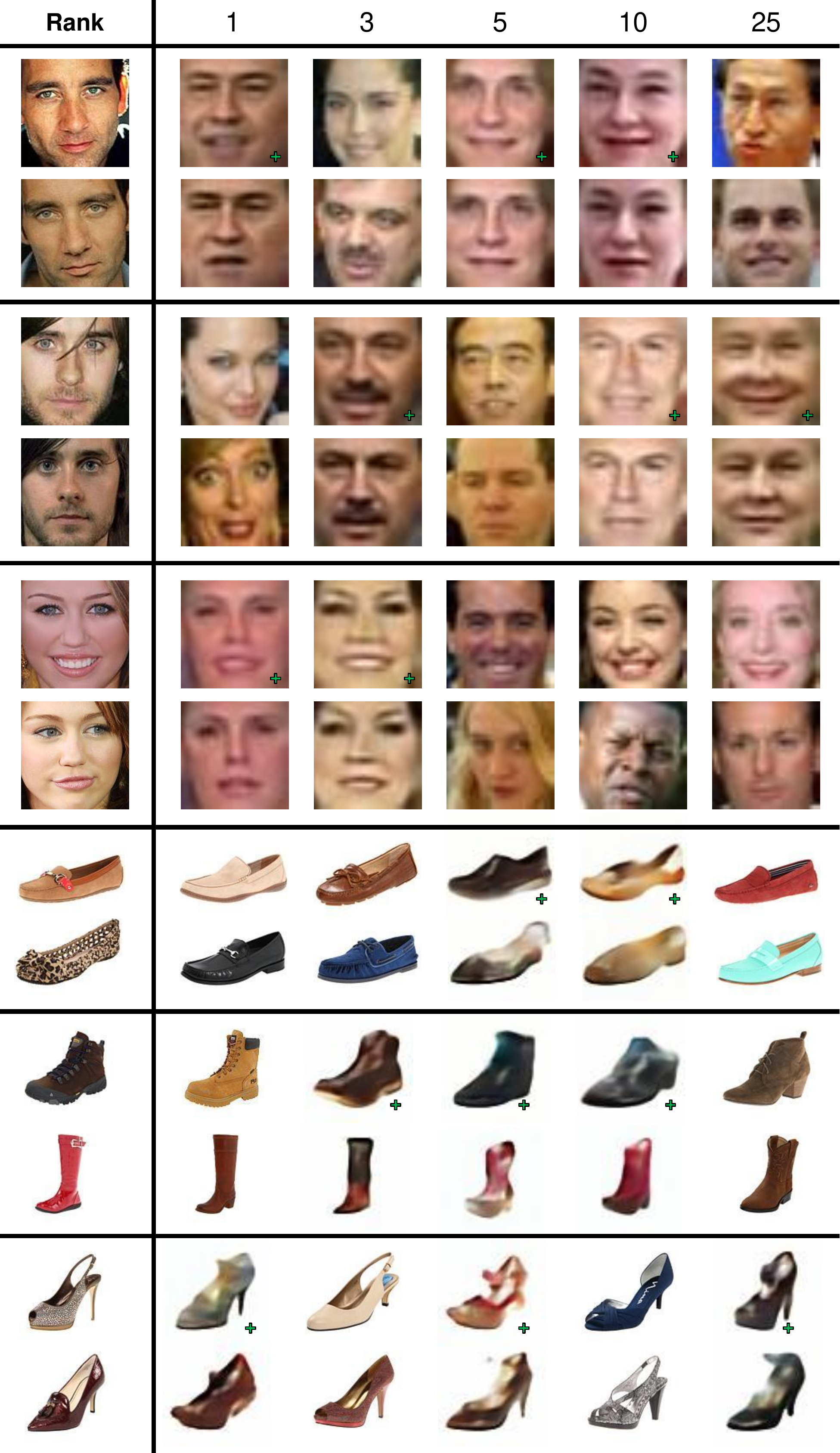}
  \caption{Examples of nearest neighbor image pairs given novel test pairs (left).  A green plus sign denotes a synthetic image pair.}
  \label{fig:neighbors}
\end{figure*}

In Figure~\ref{fig:neighbors}, we provide additional qualitative examples of the neighboring pairs given actual test pairs, expanding upon Figure~\ref{fig:qual}.  Notice that for the face images, the synthetic image pairs exhibit fine-grained differences while preserving the underlying identity, something that is valuable for learning but hard to obtain using real image pairs.


%

\end{document}